\documentclass[sigconf]{acmart}

\AtBeginDocument{%
  \providecommand\BibTeX{{%
    \normalfont B\kern-0.5em{\scshape i\kern-0.25em b}\kern-0.8em\TeX}}}
\usepackage{algorithm}
\usepackage{algorithmic}
\usepackage{booktabs}
\usepackage{balance}
\usepackage{multirow}
\usepackage{xcolor}
\usepackage[title,toc,titletoc,page]{appendix}
\usepackage{pifont}

\newlength\savewidth\newcommand\shline{\noalign{\global\savewidth\arrayrulewidth
  \global\arrayrulewidth 1pt}\hline\noalign{\global\arrayrulewidth\savewidth}}

\copyrightyear{2022}
\acmYear{2022}
\setcopyright{acmcopyright}
\acmConference[MM '22]{Proceedings of the 30th ACM
	International Conference on Multimedia}{October 10--14, 2022}{Lisboa, Portugal}
\acmBooktitle{Proceedings of the 30th ACM International Conference on Multimedia
	(MM '22), October 10--14, 2022, Lisboa, Portugal}
\acmPrice{15.00}
\acmDOI{10.1145/3503161.3548045}
\acmISBN{978-1-4503-9203-7/22/10}

\begin{document}

\title{Split-PU: Hardness-aware Training Strategy for Positive-Unlabeled Learning}

\author{Chengming Xu}
\affiliation{School of Data Science, Fudan University \country{China}}
\email{cmxu18@fudan.edu.cn}
\author{Chen Liu}
\affiliation{Department of Mathematics
Hong Kong University of Science and Technology \country{China}}
\email{cliudh@connect.ust.hk}
\author{Siqian Yang}
\author{Yabiao Wang}
\affiliation{Tencent Youtu Lab \country{China}}
\email{seasonsyang@tencent.com}
\email{caseywang@tencent.com}
\author{Shijie Zhang}
\affiliation{School of Data Science, Fudan University \country{China}}
\email{18307110401@fudan.edu.cn}
\author{Lijie Jia}
\affiliation{Shanghai Jiaotong University \country{China}}
\email{litery@sjtu.edu.cn}
\author{Yanwei Fu}
\affiliation{School of Data Science, Fudan University \country{China}}
\email{yanweifu@fudan.edu.cn}

\renewcommand{\shortauthors}{Chengming Xu et al.}

\begin{abstract}
  Positive-Unlabeled (PU)  learning   aims to learn a model with  rare positive samples and abundant unlabeled samples. Compared with classical binary classification, the task of PU learning is much more challenging due to the existence of many incompletely-annotated data instances. Since only part of the most confident positive samples are available and evidence is not enough to categorize the rest samples, many of these unlabeled data may also be the positive samples.
  Research on this topic is particularly useful and essential to many real-world tasks which demand very expensive labelling cost. For example, the recognition tasks in disease diagnosis, recommendation system and satellite image recognition may only have few positive samples that can be annotated by the experts.
  While this problem is receiving increasing attention, most of the efforts have been dedicated to the design of trustworthy risk estimators such as uPU~\cite{du2014upu} and nnPU~\cite{kiryo2017nnpu} and direct knowledge distillation, e.g., Self-PU~\cite{chen2020selfpu}. These methods mainly omit the intrinsic hardness of some unlabeled data, which can result in sub-optimal performance as a consequence of fitting the easy noisy data and not sufficiently utilizing the hard data. In this paper, we focus on improving the commonly-used nnPU~\cite{kiryo2017nnpu} with a novel training pipeline. We highlight the intrinsic difference of hardness of samples in the dataset and the proper learning strategies for easy and hard data. By considering this fact, we propose first splitting the unlabeled dataset with an early-stop strategy. The samples that have inconsistent predictions between the temporary and base model are considered as hard samples. Then the model utilizes a noise-tolerant Jensen-Shannon divergence loss for easy data; and a dual-source consistency regularization for hard data which includes a cross-consistency between student and base model for low-level features and self-consistency for high-level features and predictions, 
  respectively.
  Our method achieves much better results compared with existing methods on CIFAR10 and two medical datasets of liver cancer survival time prediction, and low blood pressure diagnosis of pregnant, individually. The experimental results validates the efficacy of our proposed method. Codes and models are available at \textcolor{blue}{\url{https://github.com/loadder/SplitPU_MM2022}}.
  
\end{abstract}



\maketitle
\section{Introduction \label{sec:intro}}
The deep models of classical recognition task, especially binary classification, has been supported by plenty of positive and negative samples, which are comprehensively annotated in general.
On the other hand, the collection of training data could not always be perfect. In many Multimedia applications, only part of the positive samples can be  annotated, while the other samples including both positive and negative ones can only be left without labels. 
These cases may happen quite often. For example, the annotator source is severely limited for tasks, such as military targets recognition from satellite images~\cite{fan2020camouflaged}, since expertise and confidential knowledge is required. In medical image analysis such as liver cancer diagnosis, the annotators can only provide confident judgement to few positive samples, while the other images without obvious symptom could also be positive but cannot be explicitly annotated.
Essentially, the Positive-Unlabeled  (PU) learning is  studied to handle these tasks.  
With a well-trained PU learning model, many real-world applications of different modalities with only positive and unlabeled data 
can thus be effectively solved, e.g. medical diagnosis, recommendation system, satellite image recognition, etc.

\begin{figure}
    \centering
    \includegraphics[scale=0.5]{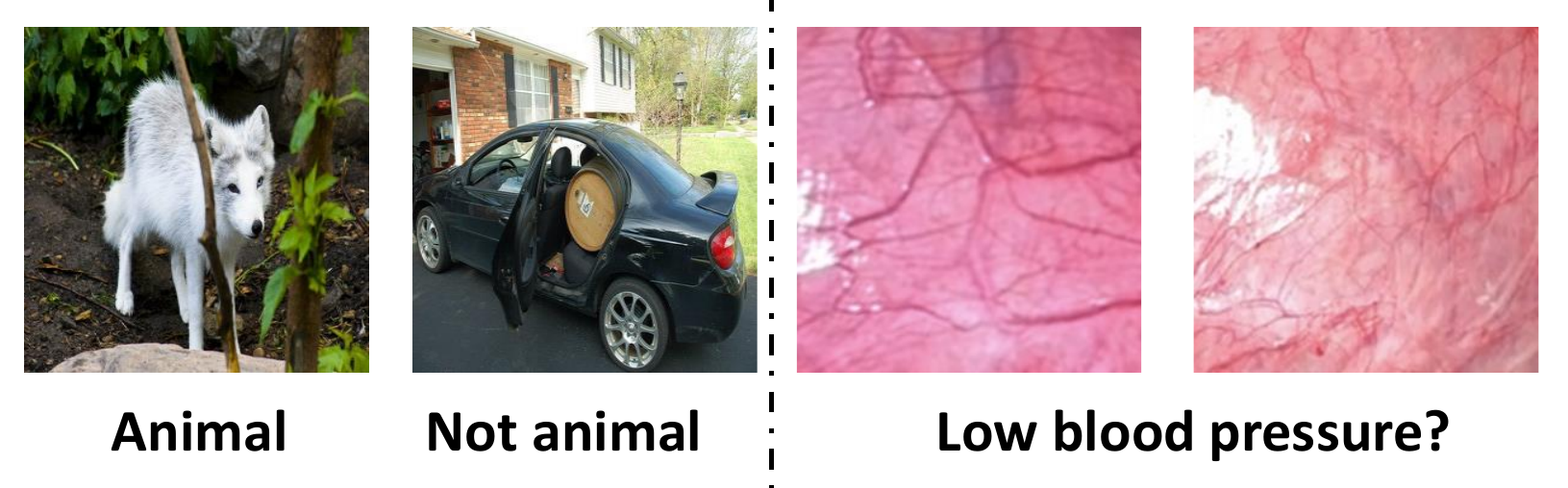}
        \vspace{-0.1in}
    \caption{ The binary classification versus PU learning. Left: The most commonly-seen visual data with distinguished categories. The data can be easily identified as an animal or not animal. Right: Data for low blood pressure diagnosis. Without sufficient expertise annotators, only part of the most confident positive samples can be recognized. Therefore the data in such an area is often taken as PU data.    \label{fig:pu}}
    \vspace{-0.15in}

\end{figure}

Due to the special structure of training set, PU learning is generally more challenging than standard supervised binary classification. Directly applying existing methods in both fully-supervised and semi-supervised learning would lead to great performance degradation. The classical PU learning methods can be traced back to confident negative sample selection\cite{liu2002partially, liu2003building} in the early stages, and focus on designing unbiased risk estimator~\cite{du2014upu, kiryo2017nnpu} recently. While some previous works like Self-PU~\cite{chen2020selfpu} and PUUPL~\cite{dorigatti2022puupl} have explored combining together the well-designed objective function and sample selection for better performance, these methods neglect an important fact in the PU data: 
\textit{Some samples are relatively much harder to learn by deep neural networks than most data}. 


Such hardness is not similar to the definition of 'easiness' depicted in Self-PU which is measured by the confidence of predictions. 
It is the intrinsic property of image data that we focus on, which could be resulted from many different reasons.
For example, for commonly-used image datasets, those images with extremely small, distorted or camouflaged objects can be relatively harder to learn than other normal one. Similarly, in medical image analysis, some samples may have severe artifacts, affecting training and inference of models. These artifacts may be caused by the errors of medical instruments, which can hardly be avoided in all cases. 

By considering this fact, we advocate that 
the unlabeled data should be further divided into easy and hard set. 
The easy data dominates the unlabeled set, being easy to fit but vulnerable to noise, while the hard data is on the opposite.
Specifically, inspired by Self-PU, we can annotate these unlabeled samples with pseudo-labels generated by a base model, e.g. a nnPU model. Then both sets contain wrongly-labeled (noisy) data and correctly-labeled (clean) data. We highlight two important points which imply that these different types of data serve as different roles in training. (1) Easy data is not totally reliable. Especially, since the easy noisy samples can be fitted by the model facilely, it is more likely for these samples than the hard noisy ones to hurt the generalization ability of the learned PU models.
Thus potentially, it may be  questionable to directly utilize 
all pseudo-labels from the easy set in a vanilla objective functions like cross entropy losses.
(2) Hard data is more difficult to deal with and requires objective functions that exclude the pseudo-labels.
As the name of hard data suggests, the model can hardly learn information from the pseudo-labels of hard data no matter whether they are clean or not. Nevertheless, the hard clean samples are essentially valuable for estimating the decision boundary. For example, as a common rule of the thumb, the hard and ambiguous training instances may also be taken as  support vectors to compute the classifier margins in support vector machine.
Therefore, it is a problem to make use of the hard data in a more effective way. Given these two points, 
the na\"ive 
selection-distillation pipeline as in Self-PU, i.e. gradually selecting confident samples and training with their pseudo-labels, 
may lead to sub-optimal performance, as all unlabeled samples are treated with the same form of objective functions. 

To this end, we propose a novel Split-PU method that employs a novel hardness-aware training strategy for positive-unlabeled learning. Our Split-PU method takes apart the easy and hard data from all PU data to deal with, respectively. Specifically, the whole method contains four steps. (1) We first  train a base teacher model of nnPU~\cite{kiryo2017nnpu} as objective function. The base model is applied to generate the initial pseudo-labels for all unlabeled data. (2) Then a temporary student model is trained via knowledge distillation from the teacher model, with recording training accuracy of each epoch. 
When the temporary model reaches the specified performance, we split out those training samples as hard data that have the inconsistent predictions between the temporary model and the base one, with the rest  as easy data.
(3) Further, different learning strategies are employed to handle the easy and hard data. For easy data, we find out that those easy noisy samples cannot be detected as effectively as the hard noisy samples by using existing noise detection methods. As an alternative, we thus for the first time introduce the Jensen-Shannon divergence loss~\cite{englesson2021djs} to optimize the PU learning model   for noisy-robust PU learning. For hard data, we develop a dual-source consistency regularization which covers low-level and high-level features together with predictions, thus fully utilizing the supervision from both the base  and the student model. We have to regularize such consistency here: \textit{a cross-consistency} between the low-level features of base model and student model, which is aimed to guide the student model with well-trained low-level features like edges and colors from base model; and \textit{a self-consistency} on the high-level features and predictions of student model between different views of each image, which is used to improve the stability of training for better visual representations. With the above training strategy, we can better handle the unlabeled data with the pseudo-labels predicted by the base model, leading to a student model with better performance. (4) Finally, we replace the base model with the student model to iterate the training algorithm for further improvement.

To validate the effectiveness of our proposed method, we conduct extensive experiments. First, one of the most widely-used visual benchmarks CIFAR10 is adopted, following the existing methods. Additionally, we adopt a MRI dataset on revival time prediction for liver cancer and a myometrioum image dataset of low blood pressure diagnosis for the pregnant to further testify the capability of our model on real-life applications. The results on these three datasets illustrate that our model can significantly improve the base model trained with nnPU and surpass the state-of-the-art methods. Moreover, when trained with fewer positive labeled data, our model still performs better than competitors which use more labeled data, which supports the argument that our model can solve the data scarcity problem to some extent.


In summary, our paper has the following contributions:
\begin{itemize}
    \item We analyze the essence of the PU data and find that those hard data could make the existing pseudo-labeling methods less effective.
    \item To take advantage of the analysis, we propose to use an early-stop splitting strategy to excavate the easy and hard data from the unlabeled set.
    \item We leverage the Jensen-Shannon divergence loss to learn the easy data, and propose a novel dual-source consistency regularization to efficiently utilize the hard unlabeled data.
    \item With the help of our proposed techniques, we consistently improve the base model among different datasets when using various amount of labeled samples.
\end{itemize}

In the following context, we first briefly introduce the related works in Sec.~\ref{sec:relate}, which includes PU learning, learning with noisy labels and consistency in deep learning. Then in Sec.~\ref{sec:method} we give the problem formulation of PU learning and our proposed method. Experiment results among multiple datasets are listed in Sec.~\ref{sec:exp}.

\section{Related Works \label{sec:relate}}

\noindent \textbf{Positive-unlabeled Learning}
Positive-unlabeled learning is a surging research topic that aims to learn patterns on a training set with only few labeled positive data and sufficient unlabeled data. The research on PU learning can be dated back to \cite{liu2002partially, liu2003building, yu2002pebl} in which reliable negative samples are selected and used to train a set of classifiers. The recent effort on PU learning with deep deep learning can be generally categorized into two buckets. One is about the design of objective function that can unbiasedly estimate the risk. uPU~\cite{du2014upu} shows that a cost-sensitive classifier can be used by reform the risk in original classification, with known class prior. nnPU~\cite{kiryo2017nnpu} points out the tendency of overfitting complex models when using uPU. As a solution they impose a non-negative constraint on the objective function of uPU, which leads to better generalization. Some other papers follow these two works to find powerful objective functions in different settings such as imbalanced data~\cite{su2021imbalance}, biased negative data~\cite{hsieh2019pubn}, data selection bias~\cite{kato2018nnpusb}, etc. Some other works focus on sample selection from the unlabeled samples based on the former methods. For example, \cite{xu2019revisiting} analyze the reasonability of only selecting positive data, Self-PU~\cite{chen2020selfpu} leverages self-paced learning to gradually update the base model with newly-learned knowledge, PUUPL~\cite{dorigatti2022puupl} strengthens the calibration of base model so that the pseudo-labels are all uncertainty-aware. Note that the concept of hardness has also been mentioned in Self-PU. However, they refer to the this concept as the reliability of pseudo-labels, which is reflected by the confidence of model predictions. Our method split the data into easy and hard set based on the intrinsic property of data, which is different with the existing ideas. Moreover, we point out that directly using cross entropy on pseudo-labeled data and PU loss on unlabeled data is inappropriate. Instead, we propose to conquer the easy and hard unlabeled data  based on noise-tolerant learning and dual-source consistency regularization respectively.
Additionally, PU learning is different with semi-supervised learning which also uses the unlabeled data as the regularizer. The unlabeled data in PU learning plays a more important role of estimating the data distribution since no labeled negative samples are available. 

\noindent \textbf{Learning with noisy labels}
Since the pseudo-labeling strategy in PU learning inevitably generates wrongly-labeled samples, i.e. noisy samples, such methods are correlated with the task of learning with noisy labels, which is aimed to train a more robust model from the noisy dataset. Methods in learning with noisy labels mainly belong to two groups: noise robust algorithm and noise detection. Noise robust algorithms target designing network architectures~\cite{chen2015webly, goldberger2016training} or objective function~\cite{zhang2018generalized, ghosh2017robust, englesson2021djs} that are robust against side effect of noisy samples. Noise detection methods involve a pipeline for selecting and processing the noisy data from the training set. Typical selection basis includes large error~\cite{shen2019learning}, incoherent gradient~\cite{chatterjee2020coherent}, inconsistency among different networks or different views of the same input~\cite{yu2019does, zhou2020robust} and so on. As for the strategies for handling noisy samples, label-correction modules are proposed in many works~\cite{li2017learning, tanaka2018joint, vahdat2017toward}. We in this paper leverage the technique in this topic to solve PU learning, which has never been studied in the previous works. Moreover, 
while directly applying these advantaged noise detection methods may also help dealing with the noisy pseudo-labels in PU learning, we will show in the experiments that these methods could easily fail when facing with some of the training data.

\noindent \textbf{Consistency Regularization} 
Recently consistency regularization has been widely used in deep learning as extra supervision besides the annotations. Specifically, contrastive learning methods~\cite{he2020moco, chen2020simclr, caron2020swav, chen2021simsiam} take as objective function the similarity among features of different views of the same image, which can be seen as the feature level consistency. Such objectives are utilized to enhance the instance discriminative ability for networks. Moreover, for semi-supervised learning, many works~\cite{sohn2020fixmatch, verma2019interpolation, jeong2019consistency} adopt prediction level consistency to increase the stability of training. Compared with these methods, our proposed dual-source consistency regularization takes into account the various properties of features and predictions, which leads to consistency with more capacity from multiple sources.

\section{Methodology \label{sec:method}}
\noindent\textbf{Problem Formulation} Suppose we have a data distribution $\mathcal{T}$ which can be represented as a joint distribution of image distribution $\mathcal{I}$ and label distribution $\mathcal{Y}$, i.e.,  $\mathcal{T}=\mathcal{I}\times \mathcal{Y}$. In PU learning, the training set $\mathcal{T}^{train}$ is composed of two parts including $\mathcal{T}^{p}$ which only contains few labeled positive samples and $\mathcal{T}^{u}$ which contains abundant unlabeled samples. The labeled positive $\mathcal{T}^{p}=\{(I_i^p, y_i^p)\}_{i=1}^{n_p}$, where $I_i^p\sim\mathcal{I}, y_i^p=1$. The unlabeled  $\mathcal{T}^{u}=\{I_i^u\}_{i=1}^{n_u}$, where $I_i^u\sim\mathcal{I}$. Our goal is to learn a network $\phi$, which can well generalize to the test set $\mathcal{T}^{test}=\{(I_i, y_i)\}_{i=1}^{n_t}$.

\begin{figure*}
    \centering
    \includegraphics[scale=0.38]{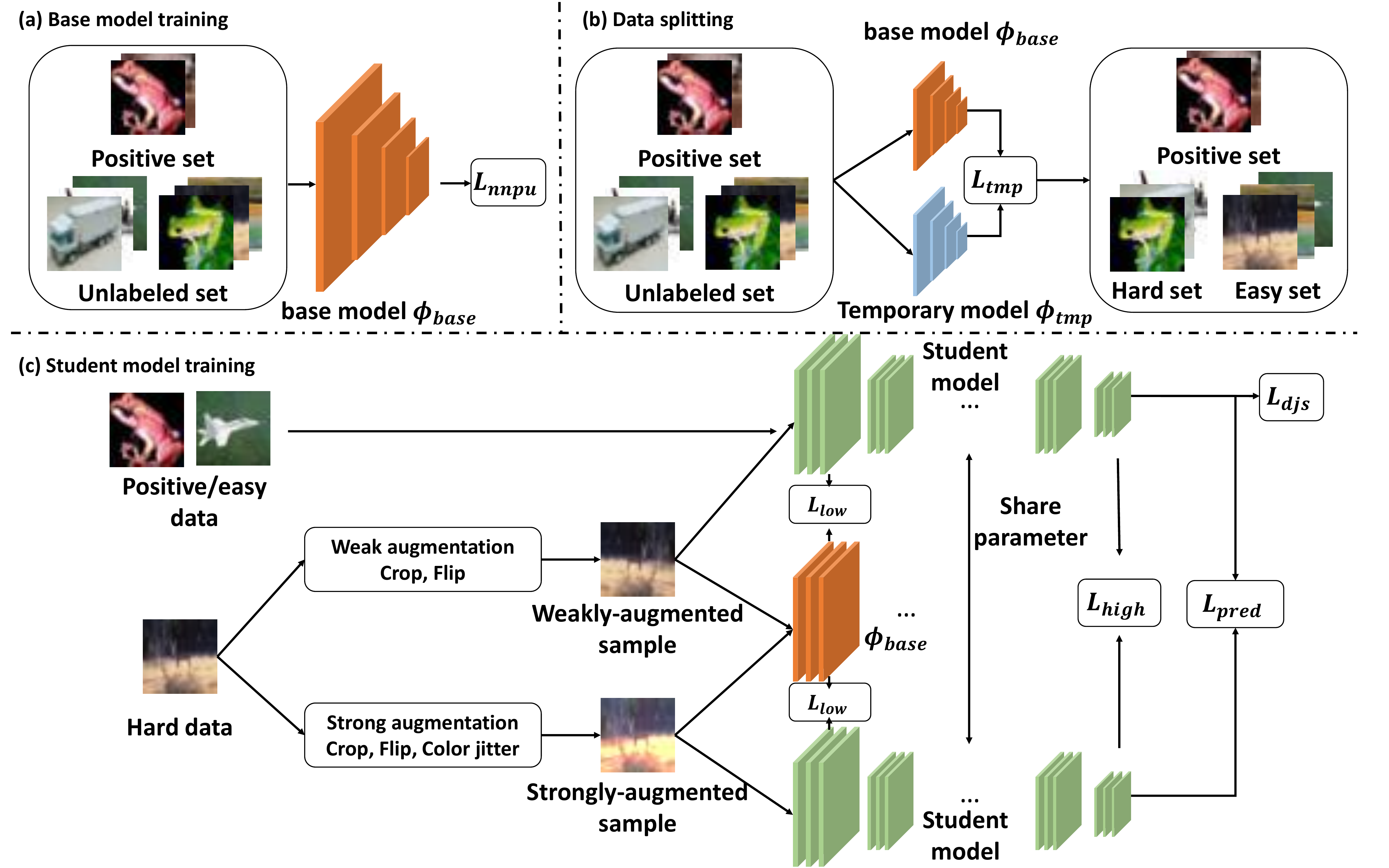}
        \vspace{-0.1in}
    \caption{Overview of our model. We first train a base teacher model using nnPU as objective function. Then the unlabeled data is split into easy and hard set via a temporary model which is trained with knowledge distillation given the base model. Then  we iterate a new training process applying different learning strategies to the easy and hard data to train the student model.}
    \label{fig:model}
        \vspace{-0.15in}
\end{figure*}

\noindent\textbf{Overview}. We propose a hardness-based splitting strategy to deal with the PU learning, which is schematically shown in Fig.~\ref{fig:model}. Concretely, we first train a base model $\phi_{base}$ with nnPU~\cite{kiryo2017nnpu} (Sec.~\ref{sec:base}), from which all of the following supervision is generated. After attaining this base model, we split the unlabeled set $\mathcal{T}^{u}$ into easy set $\mathcal{T}^{e}$ and hard set $\mathcal{T}^{h}$ by the training accuracy of a temporary student network $\phi_{tmp}$ which is trained with naive knowledge distillation from $\phi_{base}$ (Sec.~\ref{sec:split}). Then we re-train a network $\phi$ with different learning strategies on $\mathcal{T}^{easy}$ and $\mathcal{T}^{hard}$. For the easy set $\mathcal{T}^{easy}$, we utilize a noise-tolerant objective function to make use of the pseudo-labels from $\phi_{base}$ (Sec.~\ref{sec:djs}). For the hard set $\mathcal{T}^{hard}$, we adopt a dual-source consistency regularization to 
take advantage of the helpful knowledge from $\phi_{base}$ without using the pseudo-labels (Sec.~\ref{sec:dual}). Such a learning strategy can be repeated by replacing the teacher model $\phi_{base}$ with the newly-learned $\phi$ to further improve the performance not only for commonly-used visual datasets, but also for PU learning settings in real scenario.

\subsection{Preliminary: Base Model Training \label{sec:base}}

We first briefly introduce the  nnPU~\cite{kiryo2017nnpu} as our model.
The nnPU is a non-negative version of an unbiased risk estimator as following:
\begin{align}
    \label{eq:nnpu}
    L_{nnPU}&=\frac{\pi}{n_p}\sum_{i=1}^{n_p} L(\phi(I_i^p), 1) + \\
    &max \{0, \frac{1}{n_u}\sum_{i=1}^{n_u}L(\phi(I_i^u), -1)-\frac{\pi}{n_u}\sum_{i=1}^{n_u}L(\phi(I_i^p), -1)\}
\end{align}
where $\pi$ is the class prior of positive samples, i.e. $\pi=P(I_i|y_i=1)$, which is supposed to be known as~\cite{kiryo2017nnpu}. By training with $L_{nnPU}$ on $\mathcal{T}^{train}$, we can get our base model $\phi_{base}$ and correspondingly a pseudo-labeled extension of $\mathcal{T}^u$:
\begin{align}
    \hat{\mathcal{T}}^u &= \{(I_i^u, \hat{y_i}^u)\}_{i=1}^{n_u} \\
    \hat{y_i}^u &= \phi_{base}(I_i^u)
\end{align}
Note that $\hat{y_i}^u$ can be either soft-label or hard-label. In this way, $\hat{\mathcal{T}}^u$ is composed of correctly-labeled data (i.e. clean data), and wrongly-labeled data (i.e., noisy data). A straightforward way, like in Self-PU~\cite{chen2020selfpu}, to make use of $\hat{\mathcal{T}}^u$ is to train a supervised learning model based on these pseudo-labels and then iteratively rectify the   noisy or wrongly pseudo-labeled samples. However, as we will show in the experiments (see Fig.~\ref{fig:hard_data}), some training samples are intrinsically hard to learn. This implies that the models can barely receive knowledge from these samples no matter they are affiliated with correct pseudo-labels or not. Thus such a learning strategy will lead to sub-optimal performance. As a solution, we propose to use the following method to separate these hard samples from $\mathcal{T}^u$ and then learn the easy and hard data respectively.


\subsection{Early-stop Splitting Strategy \label{sec:split}}
Given the base model, we consider splitting $\mathcal{T}^u$ apart into easy and hard data, and then conquer them respectively. Exactly, we learn a new student model $\phi_{tmp}$ based on the pseudo-labels generated by $\phi_{base}$ with soft-label cross entropy as objective function:
\begin{equation}
\label{eq:tmp}
    L_{tmp} = \sum_{i=1}^{n_u}-\hat{y}_i^u log(\phi_{tmp}(I_i^u))
\end{equation}
We set a small learning rate for this optimization process to track the variety of $\phi_{tmp}$ in detail. After each epoch, we record the training accuracy. Once the accuracy is over a threshold $\tau$, we stop the training and split the unlabeled set  $\hat{\mathcal{T}}^u$ based on the prediction coherence between $\phi_{tmp}$ and $\phi_{base}$. The samples that are still not learned by $\phi_{tmp}$ are categorized as hard samples, which compose the hard set $\mathcal{T}^{hard}$. The rest are called easy set, denoted as $\mathcal{T}^{easy}$. Since the noisy samples generally have different patterns from clean samples as in  existing works~\cite{chatterjee2020coherent, arpit2017closer}, most of them would be learned more slowly than clean ones. As a consequence, $\mathcal{T}^{hard}$ can have a much larger proportion of noisy samples than $\mathcal{T}^u$, and the noise rate of $\mathcal{T}^{easy}$ is much smaller. Meanwhile, the roles of these two subsets are totally different: (1) The easy clean samples take up the most part of $\mathcal{T}^u$, from which the model learns its patterns. (2) The easy noisy samples, despite with a small sample size, can be easily fitted by the model and hurt the generalization ability. (3) The hard samples, while can hardly be learned by $\phi_{tmp}$ and have high noise rate, can regularize the model for better performance. Especially, the hard clean samples can provide data knowledge which is not contained in the easy set. Therefore, it is essential to utilize a learning strategy robust to noise on $\mathcal{T}^{easy}$ and a new supervision other than pseudo-labels on $\mathcal{T}^{hard}$. To this end, we propose the following noise-tolerant learning on easy samples and dual-source consistency regularization on hard samples.

\subsection{Noise-tolerant Learning on Easy Samples \label{sec:djs}}
As mentioned above, the easy set $\mathcal{T}^{easy}$ has two noticeable properties: (1) The noise rate is much smaller, thus most pseudo-labels are reliable. (2) The negative effect of easy noisy samples on generalization ability cannot be neglected. Directly learning the easy samples with plain objective like cross entropy will make the model be trapped by the noisy samples. A straightforward thought is if we can use noise detection methods to further cull the noisy easy samples. However, due to the fact that the slowly-learned hard noisy samples have already been taken out from $\mathcal{T}^{easy}$, most methods based on the difference of fitting speed or fitting performance between noisy and clean data could fail in our case, as we will show in experiments. As an alternative solution, we adopt a noise-robust objective function to process the easy data, which was proposed in \cite{englesson2021djs}. Concretely, for each easy sample $I_i^{easy}\in \mathcal{T}^{easy}$ and its corresponding pseudo-label $\hat{y}_i$, the following objective function is used:
\begin{equation}
    \label{eq:DJS}
    L_{easy} = \frac{(\rho D_{KL}(\phi(I_i^{easy})\|m) + (1-\rho) D_{KL}(\hat{y}_i\|m))}{-(1-\rho)log(1-\rho)}
\end{equation}
where $m=\rho \phi(I_i^{easy}) + (1-\rho) \hat{y}_i$ which is an interpolation between the pseudo-label and prediction from $\phi$, $D_{KL}$ denotes the Kullback-Leibler divergence, $\rho$ is a hyper-parameter. Accordingly, Eq.~\ref{eq:DJS} can take into account both robustness to noise and the convergence speed. As we will show in our experiments, this term is more effective than plain knowledge distillation using pseudo-labels.

\subsection{Dual-source Consistency Regularization on Hard Samples  \label{sec:dual}}
The hard set $\mathcal{T}^{hard}$ has a totally different characteristic compared with $\mathcal{T}^{easy}$ and $\hat{\mathcal{T}}^{u}$. $\mathcal{T}^{hard}$ contains much less samples than the other two sets, while its proportion of noisy samples are much larger as we have explained above. Besides, these samples can be hardly learned with the pseudo-labels, which is reflected by the low training accuracy in the splitting phase. Therefore, it is not proper to adopt the same training strategy as for the easy samples. To this end, we explore to use these data via a dual-source consistency regularization without using the pseudo-labels. Specifically, we use the following objective function to learn the hard samples:
\begin{equation}
\label{eq:hard}
    L_{hard} = L_{low} + L_{high}
\end{equation}
where $L_{low}$ is a low-level cross-consistency between $\phi_{base}$ and $\phi$, and $L_{high}$ is a high-level self-consistency of $\phi$, which will be introduced in the following.

\noindent\textbf{Low-level cross-consistency.} While the pseudo-label generated by $\phi_{base}$ is not usable due to the high noisy rate of $\mathcal{T}^{hard}$, the teacher model can still provide some guidance of visual knowledge embedded in the low-level features. According to the existing study~\cite{lee2016deep}, compared with high-level features that is correlated with the semantic category of each sample, these low-level features contain more basic and class-agnostic visual patterns, such as edges and colors. Such patterns are still reusable no matter if it leads to noisy pseudo-prediction in the $\mathcal{T}^{hard}$. Therefore, we regularize $\phi$ with the consistency between low-level features of $\phi$ and $\phi_{base}$ on the same sample. Specifically, given a hard sample $I_i^{hard}\in \mathcal{T}^{hard}$, we simultaneously extract the corresponding features from $\phi$ and $\phi_{base}$ and compute their similarity as the objective function:
\begin{equation}
    \label{eq:cross}
    L_{cross} =  \|\phi^1(I_i^{hard}) - \phi_{base}^1(I_i^{hard})\|_2
\end{equation}
where $\phi^1$ denotes the first convolutional layer of $\phi$, and the same for $\phi_{base}^1$.

\noindent\textbf{High-level self-consistency.} The guidance from $\phi_{base}$ is sub-optimal for the high-level features and the predictions. Therefore, inspired by recent research on self-supervised learning~\cite{chen2021exploring} and semi-supervised learning~\cite{sohn2020fixmatch}, we propose to utilize self-consistency on these high-level information. Specifically, for a hard sample $I_i^{hard}$, we first process it via both weak and strong augmentation pipelines $A^{weak}, A^{strong}$, which leads to two different views of this image $\hat{I}_i^{hard}=A^{weak}(I_i^{hard}), \tilde{I}_i^{hard}=A^{strong}(I_i^{hard})$. Then the high-level features are generated from the last convolutional layer of $\phi$, together with the classification prediction from which the consistency can be calculated:
\begin{align}
    \hat{X}_i &= \phi^l(\hat{I}_i^{hard}), \hat{y}_i = \phi(\hat{I}_i^{hard}) \\
    \tilde{X}_i &= \phi^l(\tilde{I}_i^{hard}), \tilde{y}_i = \phi(\tilde{I}_i^{hard}) \\
    L_{pred} &= D_{KL}(\hat{y}_i\|\tilde{y}_i)
\end{align}
After that a new predictor $f$ is introduced to transform the high-level features into another embedding space. Based on that we can regularize $\phi$ with high-level feature consistency:
\begin{equation}
    L_{feat} = \frac{1}{2}D(\hat{X}_i, f(\tilde{X}_i)) + \frac{1}{2}D(\tilde{X}_i, f(\hat{X}_i))
\end{equation}
where $D$ denotes the normalized negative cosine similarity. The high-level self-consistency can then be summarized as the combination of $L_{pred}$ and $L_{feat}$ with tunable weights $\alpha, \beta$:
\begin{equation}
    \label{eq:self}
    L_{high} = \alpha L_{pred} + \beta L_{feat}
\end{equation}
This objective function can help the model with better instance discrimination and to be more stable against disturbance on the samples. Compared with the self-supervised consistency used in Self-PU~\cite{chen2020selfpu}, our method considers the intrinsically high noise-rate of hard data, thus making better use of teacher model and leading to more effective regularization. Meanwhile, our proposed consistency does not require training multiple networks at one time so that we can reduce the computation cost.

\noindent \textbf{Remark: why not use nnPU on $\mathcal{T}^{hard}$.} One would ask if a simple solution works for the hard samples, i.e., still taking them as unlabeled samples and learning them by nnPU loss together with $\mathcal{T}^{p}$ as in Eq.~\ref{eq:nnpu}. We here provide an intuitive explanation about this problem. There are two key components that makes $L_{nnPU}$ works: (1) The $\mathcal{T}^{u}$ can well represent the data distribution. (2) We can have an accurate estimation of class prior $\pi$. However, since the the sample size of $\mathcal{T}^{hard}$ is much smaller than that of $\mathcal{T}^{u}$, there is little chance that it can provide as much information to estimate the data distribution as $\mathcal{T}^{u}$ can. On the other hand, if we think of $\mathcal{T}^{p}$ and $\mathcal{T}^{hard}$ as another training set in PU setting, then using $L_{nnPU}$ requires a new class prior $\pi'$, which cannot be easily estimated. Therefore it is inappropriate to use $L_{nnPU}$ on $\mathcal{T}^{hard}$, which will be empirically verified in Sec.~\ref{sec:exp}.

\subsection{Iterative Training}
With the above pipeline, we can get a $\phi$ that not only inherits the useful knowledge from $\phi_{base}$, but also eliminate the side-effect of noisy samples. Moreover, $\phi$ can be used as the teacher model again to train another new student model for further improvement. 

\section{Experiments \label{sec:exp}}
\begin{figure}
    \centering
    \includegraphics[scale=0.5]{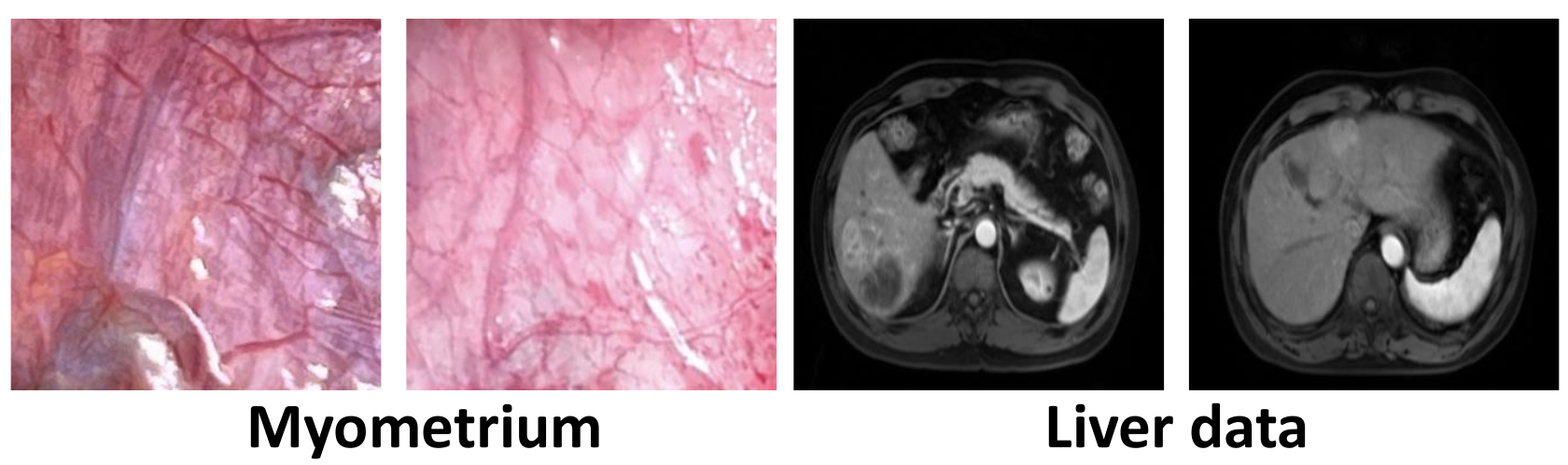}
        \vspace{-0.2in}
    \caption{Image examples in the liver cancer dataset and myometrium dataset.    \label{fig:dataset}}
        \vspace{-0.15in}
\end{figure}

\noindent\textbf{Dataset} To validate the efficacy of our model, we first follow Self-PU~\cite{chen2020selfpu} to conduct extensive experiments on CIFAR10~\cite{krizhevsky2009learning}. \textit{CIFAR10} is a basic visual dataset containing 50000 training images and 10000 testing images from 10 categories. We follow the setting in Self-PU to take 4 categories of vehicles (i.e., 'airplane', 'automobile', 'ship', 'truck') to compose the positive class and the other animal classes (i.e., 'bird', 'cat', 'deer', 'dog', 'frog', 'horse') for the negative class. In each experiment, 500/1000/3000 positive samples are randomly chosen and the rest are set as unlabeled samples.
Moreover, as shown in Fig.~\ref{fig:dataset}, we further adopt two new benchmarks of liver cancer data and myometrium data, which would be publicly released. 
The liver cancer dataset contains MRI data of 259 patients and their corresponding survival time. We split the dataset by setting samples with survival time less than 90 as positive. This leads to totally 145 positive samples, from which we randomly choose 40 as labeled samples, and 114 negative samples. For test set, we have 65 negative samples and 35 positive samples. The Myometrium data is used to diagnose the low blood pressure of the pregnant, with 327 positive images and 233 negative images in total. We split the dataset into 395 training images, 53 validation images and 112 testing images, and then randomly sample 40 positive images as labeled set.

\noindent\textbf{Implementation Details} 
For CIFAR10 we use the same network architecture as in Self-PU, i.e. a 13-layer CNN for CIFAR10. About Liver, we adopt a multi-branch ResNet18, with the detail in supplementary. For Myometrium we finetune the whole layers of an ImageNet-pretrained ResNet18.

For all datasets, we first train a base nnPU model for 50 epoches, using Adam~\cite{kingma2014adam} with learning rate of $10^{-4}$. Batch size is set to 512 for CIFAR10 and 64 for other datasets. After training the base model, we fix the splitting of labeled and unlabeled samples and then further train $\phi_{tmp}$ using SGD as optimizer with learning rate of $10^{-3}$ to avoid converging to quickly. $\phi$ is trained for 100 epoches using Adam with learning rate of $0.00005$. As for the hyper-parameters, $\tau, \rho, \alpha, \beta$ are set as 92\%, 0.7, 0.3, 0.1, respectively. We use random cropping and flipping as weak augmentation. Color jittering is adopted as strong augmentation for CIFAR10 and CutOut is adopted for the other two medical datasets.

\noindent\textbf{Metrics} For all results, we report the mean accuracies and standard deviations among 5 repetitions of each experiment.

\noindent\textbf{Competitors} We choose several former methods in PU learning as our competitors including uPU~\cite{du2014upu}, nnPU~\cite{kiryo2017nnpu}, DAN~\cite{liu2019discriminative}, Self-PU~\cite{chen2020selfpu}, P$^3$MIX-C~\cite{li2021your}, PUUPL~\cite{dorigatti2022puupl}.

\begin{table}[t]
 \centering
 {
  \begin{tabular}{ l|ccc}
  \hline
 \multirow{2}{*}{Model} & \multicolumn{3}{c}{CIFAR10} \tabularnewline
 \cline{2-4}
& 500 & 1000  & 3000\\

   \shline
     uPU~\cite{du2014upu} & --- & 88.00$\pm$0.62 & ---\tabularnewline
     \cline{2-4}
     nnPU~\cite{kiryo2017nnpu} & --- & 88.60$\pm$0.40 & ---\tabularnewline
     \cline{2-4}
     nnPU$^{*}$ & 87.22$\pm$0.46 & 89.02$\pm$0.39 & 90.53$\pm$0.36\tabularnewline
     \cline{2-4}
     \hline
     Self-PU~\cite{chen2020selfpu} & --- & 89.68$\pm$0.22 & 90.77$\pm$0.21\tabularnewline
     \cline{2-4}
     PUbN~\cite{hsieh2019pubn} & --- & 89.30$\pm$0.57 & ---\tabularnewline
     \cline{2-4}
     P$^3$MIX-C~\cite{li2021your} & --- & 87.90$\pm$0.50 & ---\tabularnewline
     \cline{2-4}
     VPU~\cite{chen2020variational} & --- & 85.06$\pm$0.55 & 87.50$\pm$1.05\tabularnewline
     \cline{2-4}
     DAN~\cite{liu2019discriminative} & --- & --- & 89.70$\pm$0.40\tabularnewline
     \cline{2-4}
     PUUPL~\cite{dorigatti2022puupl} & --- & 89.84$\pm$0.13 & 91.37$\pm$0.05\tabularnewline
     \cline{2-4}
     \hline
     Ours & \textbf{89.18$\pm$0.12} & \textbf{90.51$\pm$0.10} & \textbf{92.51$\pm$0.10}\tabularnewline
     \cline{2-4}
     \hline
  \end{tabular}
 }
 \vspace{0.1in}
\caption{\label{tab:cifar} Test accuracies and standard deviation on CIFAR10 with three positive size, {*} denotes results reproduced by us.}
\end{table}

\begin{table}[t]
 \centering
 {
  \begin{tabular}{ l|cc}
  \hline
 Model & Myometrium & Liver \tabularnewline
 \cline{2-3}

   \shline
     uPU~\cite{du2014upu} & 60.18$\pm$2.33 & 40.40$\pm$5.50\tabularnewline
     \cline{2-3}
     nnPU~\cite{kiryo2017nnpu} & 62.32$\pm$2.71 & 56.60$\pm$5.90 \tabularnewline
     \cline{2-3}
     \hline
     Self-PU~\cite{chen2020selfpu} & 59.11$\pm$1.16 & 55.60$\pm$8.17\tabularnewline
     \cline{2-3}
     \hline
     Ours & \textbf{65.89$\pm$2.13} & \textbf{63.80$\pm$1.64} \tabularnewline
     \cline{2-3}
     \hline
     Supervised & 72.32 & 68.00 \tabularnewline
     \cline{2-3}
     \hline
  \end{tabular}
 }
 \vspace{0.1in}
\caption{\label{tab:medical} Test accuracies and standard deviation on Myometrium and Liver. All results of competitors are reproduced by us.}
\end{table}

\subsection{Main Results}
We compare our model with the competitors in Tab.~\ref{tab:cifar} and Tab.~\ref{tab:medical}. Note that for nnPU on CIFAR10 we both report the results in the original paper and the one reproduced by us.

\noindent\textbf{CIFAR10 result.} For CIFAR10, we train our model among three different positive sample size. When $n_p=500$, the test accuracy of our model is 1.96\% better than that of nnPU, and comparable with nnPU model trained with 1000 labeled positive samples. This means that support nnPU with our proposed method is able to solve the problem of data scarcity for positive samples. When $n_p=1000$, which is the most common setting, our proposed model attains 1.49\% accuracy increasement against nnPU and performs 0.67\% better than the best counterpart PUUPL. When $n_p=3000$, the improvement on nnPU of our model reaches 1.6\%, leading to a much better result than all of the competitors. Such an improvement is consistently significant among these three settings, which validates the effectiveness of our proposed method and shows the potential that our model can be used as a fixed suppplement for nnPU. As for the standard deviation, our model is on par with the competitors, which shows our method is stable enough to be safely utilized.

\noindent\textbf{Myometrium and Liver results.} We report results of different methods and one oracle model trained with whole labeled training set together on the two medical datasets in Tab.~\ref{tab:medical}. For Myometrioum, our model is better than nnPU by 3.75\% with comparable standard deviation. For Liver, our model improves nnPU by 7.2\% with remarkably more stability. It is noticeable that Self-PU is worse than nnPU on these datasets. We suspect this is due to the poor performance of nnPU leads to unreliable pseudo-labels, thus increasing the size of clean noisy samples. In this way, since Self-PU cannot well handle such data, the performance could be damaged. 

\subsection{Ablation Study}
To further verify the efficacy of our contributions, we conduct several ablation studies on CIFAR10, including each part of our model and the choice of hyper-parameters.

\begin{figure*}
    \centering
    \includegraphics[scale=0.38]{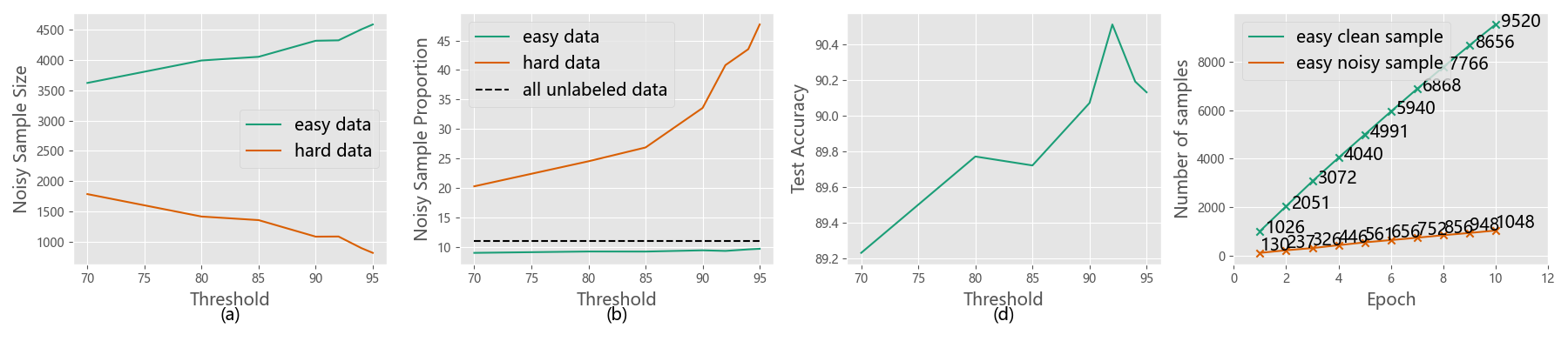}
        \vspace{-0.1in}
    \caption{(a) Size of noisy samples in easy and hard set with different $\tau$. (b) Proportion of noisy samples in easy and hard set with different $\tau$. (c) Test accuracy when trained with the easy and hard set that are generated with different $\tau$. (d) Number of clean and noisy samples that are selected from the clean set by SPR in each epoch.}
    \label{fig:early_stop}
        \vspace{-0.15in}
\end{figure*}

\begin{table}[t]
 \centering
 {
  \begin{tabular}{ l|ccc}
  \hline
  \multirow{2}{*}{Early stop} & \multicolumn{3}{c}{Acc.} \tabularnewline
 \cline{2-4}
 & 500 & 1000  & 3000\\
   \shline
     w/o & 88.57 & 89.86 & 91.73\tabularnewline
     w  & 89.28 & 90.51  & 92.51\tabularnewline
     \hline
  \end{tabular}
 }
 \vspace{0.1in}
\caption{\label{tab:early_stop} Test accuracies on CIFAR10 of models trained with and without early stop splitting.}
\end{table}

\begin{table}[t]
 \centering
 {
  \begin{tabular}{ l|c|ccc}
  \hline
  \multirow{2}{*}{Loss} & \multirow{2}{*}{label type} & \multicolumn{3}{c}{Acc.} \tabularnewline
 \cline{3-5}
 & & 500 & 1000  & 3000\\
   \shline
     \multirow{2}{*}{CE} & hard & 87.51 & 89.02 & 91.55\tabularnewline
      & soft & 88.15 & 89.48 & 92.06\tabularnewline
     \hline
     \multirow{2}{*}{DJS} & hard & 88.61 & 89.58 & 92.11\tabularnewline
      & soft & 89.28  & 90.51  & 92.51\tabularnewline
     \hline
  \end{tabular}
 }
 \vspace{0.1in}
\caption{\label{tab:easy_loss} Test accuracies on CIFAR10 of our model trained with soft cross entropy, hard cross entropy, soft Jensen-Shannon Divergence and hard Jensen-Shannon Divergence.}
\vspace{-0.1in}
\end{table}

\begin{table}[t]
 \centering
 {
  \begin{tabular}{ l|ccc}
  \hline
 \multirow{2}{*}{Loss} & \multicolumn{3}{c}{Acc.} \tabularnewline
 \cline{2-4}
 & 500 & 1000  & 3000\\
   \shline
     no & 88.52 & 89.71 & 91.46\tabularnewline
     nnPU & 87.91 & 89.43  & 90.66\tabularnewline
     self-consistency  & 89.03 & 90.09  & 92.11\tabularnewline
     cross consistency & 89.17 &  90.42 & 92.27\tabularnewline
     dual-source consistency & 89.28 & 90.51  & 92.51\tabularnewline
     \hline
  \end{tabular}
 }
 \vspace{0.1in}
\caption{\label{tab:hard_loss} Test accuracies on CIFAR10 of our model trained with different losses on hard data.}
\end{table}

\noindent\textbf{What can early-stop splitting do?} First of all, we would like to show the effectiveness of the early-stop splitting strategy. To do so, we conduct an oracle experiment where all training set labels are available for indicating which samples are wrongly labeled by $\phi_{base}$. Then we gradually increase the accuracy threshold $\tau$ and visualize the sample size and proportion of noisy samples that are categorized as easy and hard ones respectively. The results are shown in Fig.~\ref{fig:early_stop}. We can find that (1) As $\tau$ increases, less samples can be filtered out from $\mathcal{T}^{u}$, but the noise rate of $\mathcal{T}^{hard}$ gets larger. (2) When $\tau$ is relatively small as 70\%, the proportion of noisy sample in $\mathcal{T}^{hard}$ is twice as larger than that of $\mathcal{T}^{u}$, which coincides with the common sense that noisy samples are generally learned more slowly than clean samples. Also, such a strategy can effectively reduce the noise rate of $\mathcal{T}^{easy}$. (3) There is a balance between the threshold and the test accuracy. The benefit from increasing $\tau$ saturates at about 92\%, then the test accuracy gradually decreases. This is reasonable since larger $\tau$ means smaller hard set, which leads to a less representative hard set. (4) As shown in \ref{tab:early_stop}, performance is worse when early-stop strategy is not used , i.e., $L_{easy}$ is applied to all unlabeled data, which indicates the efficacy of this strategy.

\noindent\textbf{Can noise detection methods fit in our case?} As analyzed in Sec.~\ref{sec:djs}, noise detection methods may not be proper for further detecting noisy samples in $\mathcal{T}^{easy}$. To verify this, we adopt two state-of-the-art methods in noisy label learning, SPR~\cite{wang2022scalable} and CGH~\cite{chatterjee2020coherent}. SPR is based on penalized regression to build a regularization path, which is used to detect the most confident noisy samples. Fig.~\ref{fig:early_stop}(d) shows the result using SPR. With about 9\% noisy samples in $\mathcal{T}^{easy}$, SPR can only filter out a set with noise rate of 10\% in each epoch, which is even worse than uniform sampling. The result of CGH which is provided in the supplementary is also unsatisfactory, failing to detect enough noisy samples. Considering these results, we speculate that the easy set is too challenging for these noise detection methods to work.

\noindent\textbf{Choices of objective functions for easy data.} In Tab.~\ref{tab:easy_loss} we compare the noise-tolerant DJS loss $L_{easy}$ with cross entropy, using both hard and soft labels. The results demonstrate that: (1) DJS loss is more effective than cross entropy. When using soft label, DJS loss is better than cross entropy by 1.13\%, 1.03\% and 0.45\% when using 500, 1000 and 3000 positive samples. (2) Using soft label is better than hard label no matter which objective function is utilized in training. We think this is due to the base model is calibrated enough. Therefore transferring low-confidence soft labels to hard labels could hurt the model training. 

\begin{table}[t]
 \centering
 {
  \begin{tabular}{ l|ccc}
  \hline
 \multirow{2}{*}{\#Iter} & \multicolumn{3}{c}{Acc.} \tabularnewline
 \cline{2-4}
 & 500 & 1000  & 3000\\
   \shline
     1 & 88.86 & 90.07 & 92.02\tabularnewline
     2 & 89.28 & 90.51 & 92.51\tabularnewline
     3 & 89.19 & 90.55 & 92.64\tabularnewline
     4 & 89.12 & 90.49 & 92.93\tabularnewline
     \hline
  \end{tabular}
 }
 \vspace{0.1in}
\caption{\label{tab:iter} Test accuracies on CIFAR10 of our model trained with different number of iterations.}
\end{table}

\noindent\textbf{Choices of objective functions for hard data.} In Tab.~\ref{tab:hard_loss} we compare several choices to optimize hard data, including (1) not using hard data. (2) nnPU loss on hard data and positive labeled data. (3) cross consistency as in Eq.~\ref{eq:cross}. (4) self-consistency as in Eq.~\ref{eq:self}. (5) dual-source consistency as in Eq.~\ref{eq:hard}. The results reflect several important points: (1) Either forms of consistency can improve the model. Specifically, the model using self-consistency is better than the model without using hard data by 0.51\%, 0.32\% and 0.65\% among three settings, and the model using cross consistency is better by 0.65\%, 0.71\% and 0.81\%. (2) Using $L_{nnPU}$ worsens the performance by 0.61\%, 0.32\% and 0.80\%, which is compatible with our analysis in Sec.~\ref{sec:dual}. (3) Cross consistency is more effective than self-consistency, which is reasonable because cross consistency can guide the model to learn better low-level feature. Based on such features, the model can then learn to build useful high-level features easily.

\noindent\textbf{Relationship between test accuracy and number of iterations.} Since our method utilizes an iterative training strategy, we investigate how the number of iterations could affect the final result. Tab.~\ref{tab:iter} shows that while training for two iterations is much better than single iteration, i.e. 0.44\% improvement, repeating the algorithm more times does not significantly boost the performance when using 500 or 1000 positive samples. Under the setting of 3000 positive samples, training 4 iterations leads to the best result. The drawback of using more iterations is the longer training time, which could make the algorithm less practical and not comparable with the existing methods. Also, since the variation caused by the randomness can stack during each iteration, the final result with high iteration number would be more unstable.

\begin{table}[]
 \centering
 {
  \begin{tabular}{ l|ccc}
  \hline
 \multirow{2}{*}{Usage} & \multicolumn{3}{c}{Acc.} \tabularnewline
 \cline{2-4}
 & 500 & 1000  & 3000\\
   \shline
     all data & 88.96 & 90.49 & 92.25\tabularnewline
     hard data & 89.28 & 90.51 & 92.51\tabularnewline
     \hline
  \end{tabular}
 }
 \vspace{0.1in}
\caption{\label{tab:consis} Test accuracies on CIFAR10 of our model trained with dual-source consistency applied to all unlabeled data or only hard data.}
\end{table}

\noindent\textbf{Usage of consistency regularization} As a straightforward thought, our proposed dual-source consistency regularization can also be applied to easy data. We testify this idea with results in Tab.~\ref{tab:consis}. By applying dual-source consistency on all unlabeled data, the model gets accuracy of 88.96\%, 90.49\% and 92.25\%, receiving no marginal advancement. We think the reason is that since most pseudo-labels of easy data is reliable and easy to fit, the supervision with regard to semantic labels would be stronger than other forms of guidance. Therefore, using DJS loss is enough for the easy data.

\noindent\textbf{What kind of data is selected by early-stop splitting?} We visualize some positive and negative data from easy and hard set of three datasets in Fig.~\ref{fig:hard_data}. We can find that the easy data is discriminative enough, with obvious patterns and symptoms. On the other side, the hard samples are generally much harder to distinguish. For example the hard negative sample of CIFAR10 is actually an aeroplane but also looks like a bird, and the hard positive sample in Myometrium does not have the same symptoms like apparent blood vessels with dark red color.

\section{Conclusion}
In this paper we explore the task of PU learning. The problem of intrinsic hardness of some unlabeled samples is highlighted which may affect the existing methods' performance. We analyze the different role of easy and hard data in training and propose a novel training strategy as a solution to this problem. Our model includes an early-stop splitting strategy, a noise-tolerant learning strategy for easy data and a dual-source consistency regularization for hard data. We achieve remarkable results on CIFAR10 and two medical datasets, which shows the strength of our method as a promising choice for solving PU learning both in scientific research and in real-life applications.

\bibliographystyle{ACM-Reference-Format}
\balance
\bibliography{acmart}

\begin{appendices}
\section{Experimental Detail \label{sec:exp}}

\subsection{Detailed Information for Liver dataset}
Each sample in the Liver dataset denotes a patient with liver cancer. The data consists of several modalities of MRI images from all patient, including T1WI, T1CE1, T1IP and T2WI. We simply adopt T1CE1 as the input modality of our experiment and leave the full usage of all modalities as future works. the model reads all of the slices of each sample with Pydicom and resize them to $256\times 256$ without additional normalization. The labels are transformed from the survival time of each patient, which ranges from 0 to 120.

\begin{figure}
    \centering
    \includegraphics[scale=0.15]{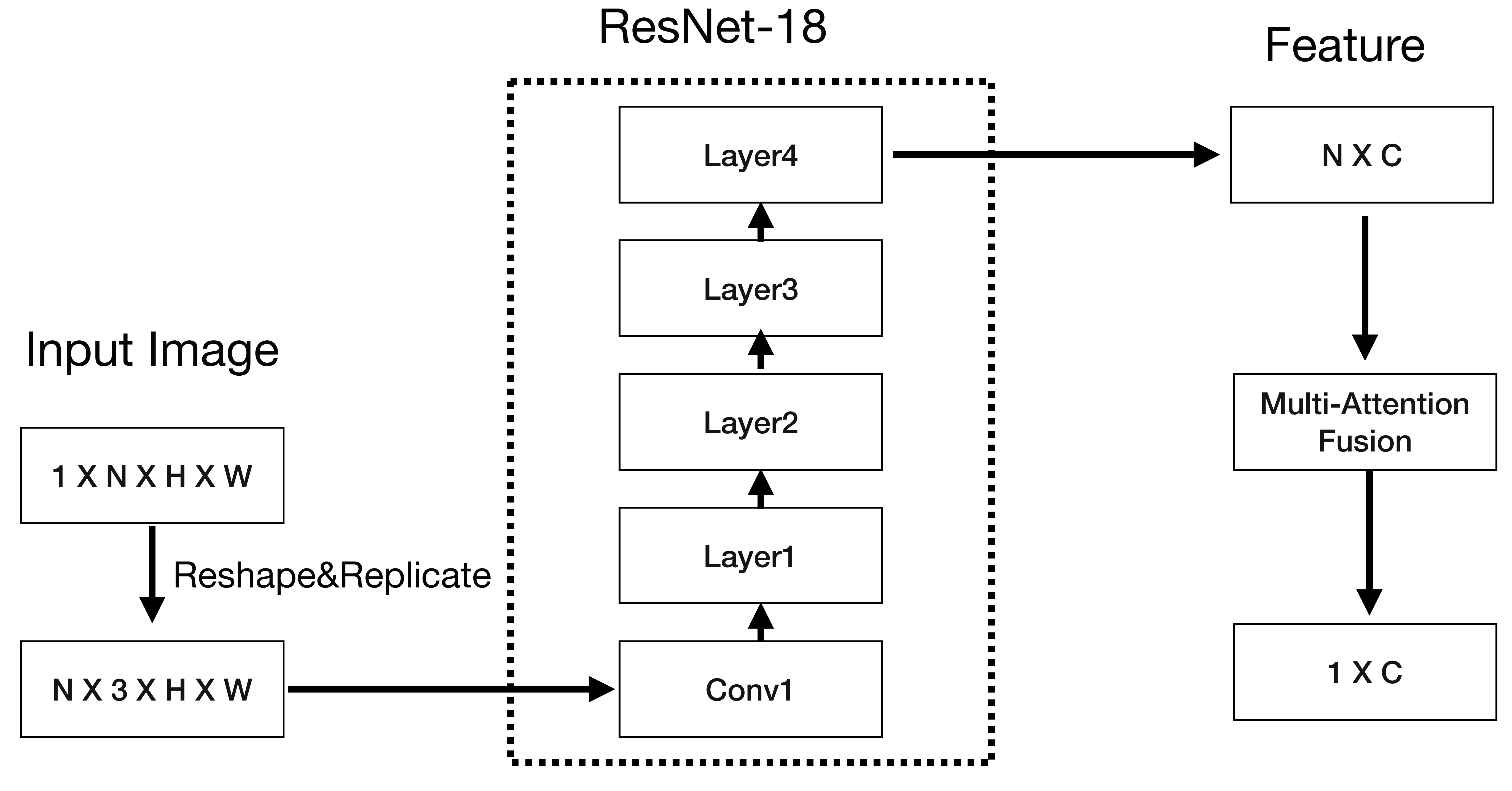}
    \caption{Detailed architecture of our network for Liver dataset. N, C, H, W denotes number of slices, number of channels, image height and image width respectively.}
    \label{fig:liver_resnet}
\end{figure}

\subsection{Detailed Architecture for Liver dataset} 
As we have mentioned in our main context, since each liver cancer sample is composed of several slices, we adopt a special architecture to deal with such data. Essentially, as shown in Fig.~\ref{fig:liver_resnet}, the basic ResNet18 is used to extract features for all slices. The network weights are shared among different slices. Then for slices from one sample, a multi-head attention module is used to aggregate these features. Then a pooling layer and a classifier are used, as the same in original ResNet18.

\section{Additional Experiment Results}
\begin{figure}
    \centering
    \includegraphics[scale=0.5]{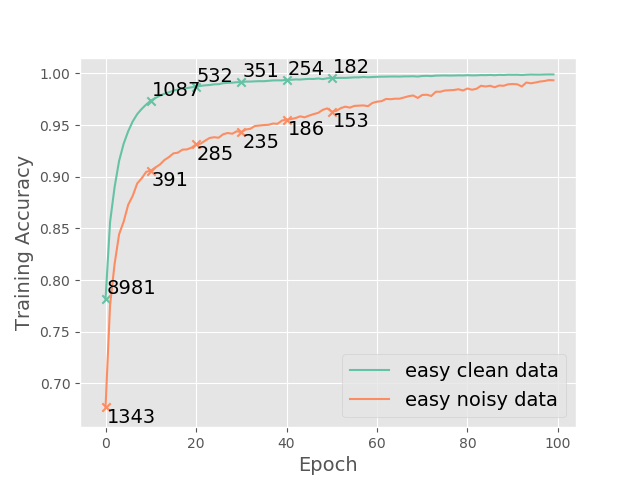}
    \caption{Training accuracy of clean and noisy data respectively in easy set.}
    \label{fig:acc}
\end{figure}

\begin{figure}
    \centering
    \includegraphics[scale=0.35]{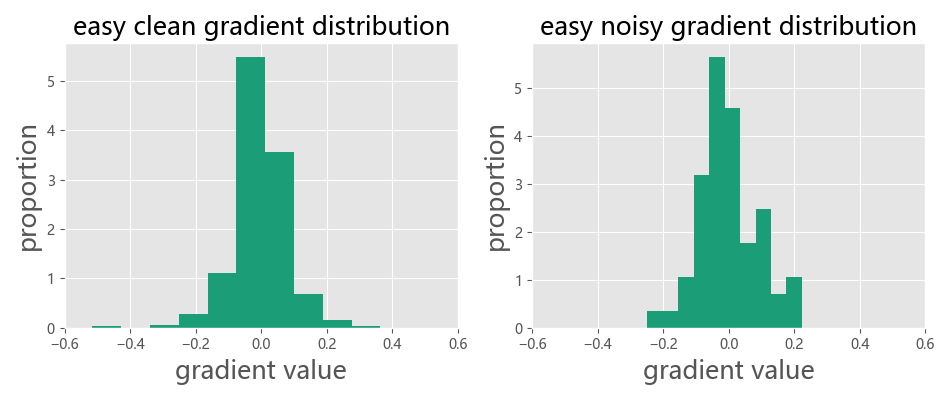}
    \caption{Gradient visualization of both clean and noisy samples on CIFAR10.}
    \label{fig:grad}
\end{figure}

\subsection{Results on noise detection methods} To support our claim in the main context that noise detection methods may not be proper for further detecting noisy samples in $\mathcal{T}^{easy}$, we first plot the training accuracy of noisy and clean samples in the easy set. As depicted in Fig.~\ref{fig:acc}, while the training accuracy of clean samples is higher than that of noisy ones for each epoch, the actual size of wrongly-predicted clean samples is also much higher than noisy set. The proportion of wrongly-predicted noisy samples among all samples that are wrongly predicted is much smaller than that in the original unlabeled data as. This means these easy noisy samples do not enjoy the slow-learning property to the same extend as the hard noisy ones. Moreover, the result of CGH~\cite{chatterjee2020coherent} is shown in Fig.~\ref{fig:grad}, in which we visualize the gradient of the first weight in the first convolutional layer of the whole network with regard to both easy noisy samples and easy clean samples. We can find that both the scale and distribution are similar between these two kinds of data, which means the coherent gradient hypothesis also does not hold so well on the easy data.

\begin{table}[]
 \centering
 {
  \begin{tabular}{ l|ccc}
  \hline
 \multirow{2}{*}{$\rho$} & \multicolumn{3}{c}{Acc.} \tabularnewline
 \cline{2-4}
 & 500 & 1000  & 3000\\
   \shline
     0.3  & 88.26 & 89.47 & 91.75 \tabularnewline
     0.4  & 88.68 & 89.51 & 92.03 \tabularnewline
     0.5  & 88.55 & 90.02 & 92.06 \tabularnewline
     0.6  & 88.92 & 90.24 & 92.03 \tabularnewline
     0.7  & 89.28 & 90.51 & 92.51 \tabularnewline
     0.8  & 89.18 & 90.31 & 92.13 \tabularnewline
     0.9  & 89.16 & 90.23 & 92.17 \tabularnewline
     \hline
  \end{tabular}
 }
 \vspace{0.1in}
\caption{\label{tab:rho} Test accuracies on CIFAR10 of our model trained with different $\rho$ in Eq.~(6) in the main context.}
\end{table}

\subsection{Choices of $\rho$} 
We compare different $\rho$ in Eq.~(6) in the main context, with the results in Tab.~\ref{tab:rho}. As mentioned in the original paper~\cite{englesson2021djs}, smaller $\rho$ leads to faster convergence and less robustness, which is roughly consistent with our results. When $\rho=0.3$, the model gets 88.26\%, 89.47\% and 91.75\%, which is the worst among all $\rho$ and almost similar to the results when using soft cross entropy. As $\rho$ gets larger, the performance gets better and saturates at $\rho=0.7$.

\begin{table}[t]
 \centering
 {
  \begin{tabular}{ l|ccc}
  \hline
  \multirow{2}{*}{Model} & \multicolumn{3}{c}{Acc.} \tabularnewline
 \cline{2-4}
 & 500 & 1000  & 3000\\
   \shline
     uPU & 85.98$\pm$1.09 & 88.22$\pm$0.28 & 89.83$\pm$0.42\tabularnewline
     uPU+Split-PU  & 87.44$\pm$0.18 & 90.57$\pm$0.26  & 91.51$\pm$0.13\tabularnewline
     \hline
  \end{tabular}
 }
 \vspace{0.1in}
\caption{\label{tab:upu} Test accuracies on CIFAR10 of models trained with and without early stop splitting.}
\end{table}

\subsection{Choices of $\alpha$ and $\beta$}
We conduct the comparison on $\alpha$ and $\beta$ with the same setting as in the main paper, i.e. 500/1000/3000 positive labeled samples on CIFAR10. As shown in Tab.~\ref{tab:alpha} and Tab.~\ref{tab:beta}, the performance of our model among different hyper-parameter settings is relatively robust. Meanwhile, the trend of performance is generally consistent with few exceptions. For example, $\alpha=0.5$ or $\beta=0.2$ lead to the highest accuracy when using 3000 positive labeled samples. We advocate that these two hyper-parameters control the balance among different loss terms adopted by our method. In particular, larger $\alpha$ and $\beta$ lead to the scale of $L_{pred}$ and $L_{feat}$, as depicted in Eq.~(11) and Eq.~(12), larger than that of $L_{easy}$. The strength of annotations is thus weakened in this way. Consequently, the model becomes poorer at learning label information from the dataset, which leads to worse performance. On the other hand, smaller $\alpha$ makes the effect of prediction consistency negligible, hence resulting in lower accuracy. We will add these results to our paper.

\begin{table}[]
 \centering
 {
  \begin{tabular}{ l|ccc}
  \hline
 \multirow{2}{*}{$\alpha$} & \multicolumn{3}{c}{Acc.} \tabularnewline
 \cline{2-4}
 & 500 & 1000  & 3000\\
   \shline
   0.1  & 87.68 & 89.87 & 90.84 \tabularnewline
   0.2  & 88.05 & 90.08 & 92.05 \tabularnewline
   0.3  & 89.18 & 90.51 & 92.51 \tabularnewline
     0.4  & 88.61 & 90.03 & 92.27 \tabularnewline
     0.5  & 88.52 & 90.08 & 92.55 \tabularnewline
     0.6  & 88.48 & 89.93 & 92.41 \tabularnewline
     0.7  & 88.60 & 90.09 & 92.13 \tabularnewline
     0.8  & 88.68 & 89.56 & 91.17 \tabularnewline
     0.9  & 88.30 & 89.30 & 90.53 \tabularnewline
     \hline
  \end{tabular}
 }
 \vspace{0.1in}
\caption{\label{tab:alpha} Test accuracies on CIFAR10 of our model trained with different $\alpha$ in Eq.~(13) in the main context.}
\end{table}

\begin{table}[]
 \centering
 {
  \begin{tabular}{ l|ccc}
  \hline
 \multirow{2}{*}{$\beta$} & \multicolumn{3}{c}{Acc.} \tabularnewline
 \cline{2-4}
 & 500 & 1000  & 3000\\
   \shline
   0.1  & 89.18 & 90.51 & 92.51 \tabularnewline
   0.2  & 88.96 & 90.31 & 92.63 \tabularnewline
     0.3  & 89.02 & 90.39 & 92.37 \tabularnewline
     0.4  & 88.67 & 90.11 & 92.43 \tabularnewline
     0.5  & 88.84 & 89.87 & 92.14 \tabularnewline
     0.6  & 88.81 & 90.09 & 92.37 \tabularnewline
     0.7  & 89.07 & 90.08 & 91.71 \tabularnewline
     0.8  & 88.48 & 89.14 & 91.24 \tabularnewline
     0.9  & 88.26 & 89.49 & 91.54 \tabularnewline
     \hline
  \end{tabular}
 }
 \vspace{0.1in}
\caption{\label{tab:beta} Test accuracies on CIFAR10 of our model trained with different $\beta$ in Eq.~(13) in the main context.}
\end{table}

\subsection{Application of Split-PU on other base models}
To further verify the efficacy of our proposed method, we use another popular objective function uPU~\cite{du2014upu} as our base model, i.e. replacing $L_{nnPU}$ with the following $L_{uPU}$ when training $\phi_{base}$:
\begin{align}
    L_{uPU}&=\frac{\pi}{n_p}\sum_{i=1}^{n_p} L(\phi(I_i^p), 1) + \\
    &\frac{1}{n_u}\sum_{i=1}^{n_u}L(\phi(I_i^u), -1)-\frac{\pi}{n_u}\sum_{i=1}^{n_u}L(\phi(I_i^p), -1)
\end{align}
The results on CIFAR10 are shown in Tab.~\ref{tab:upu}, which indicates that the improvement of our Split-PU on both nnPU and uPU is consistent. It is noticeable that when using 1000 positive labeled samples, using uPU with Split-PU is even better than using nnPU with Split-PU, receiving accuracy of 90.57\% with comparable standard deviation.

\begin{figure}
    \centering
    \includegraphics[scale=0.42]{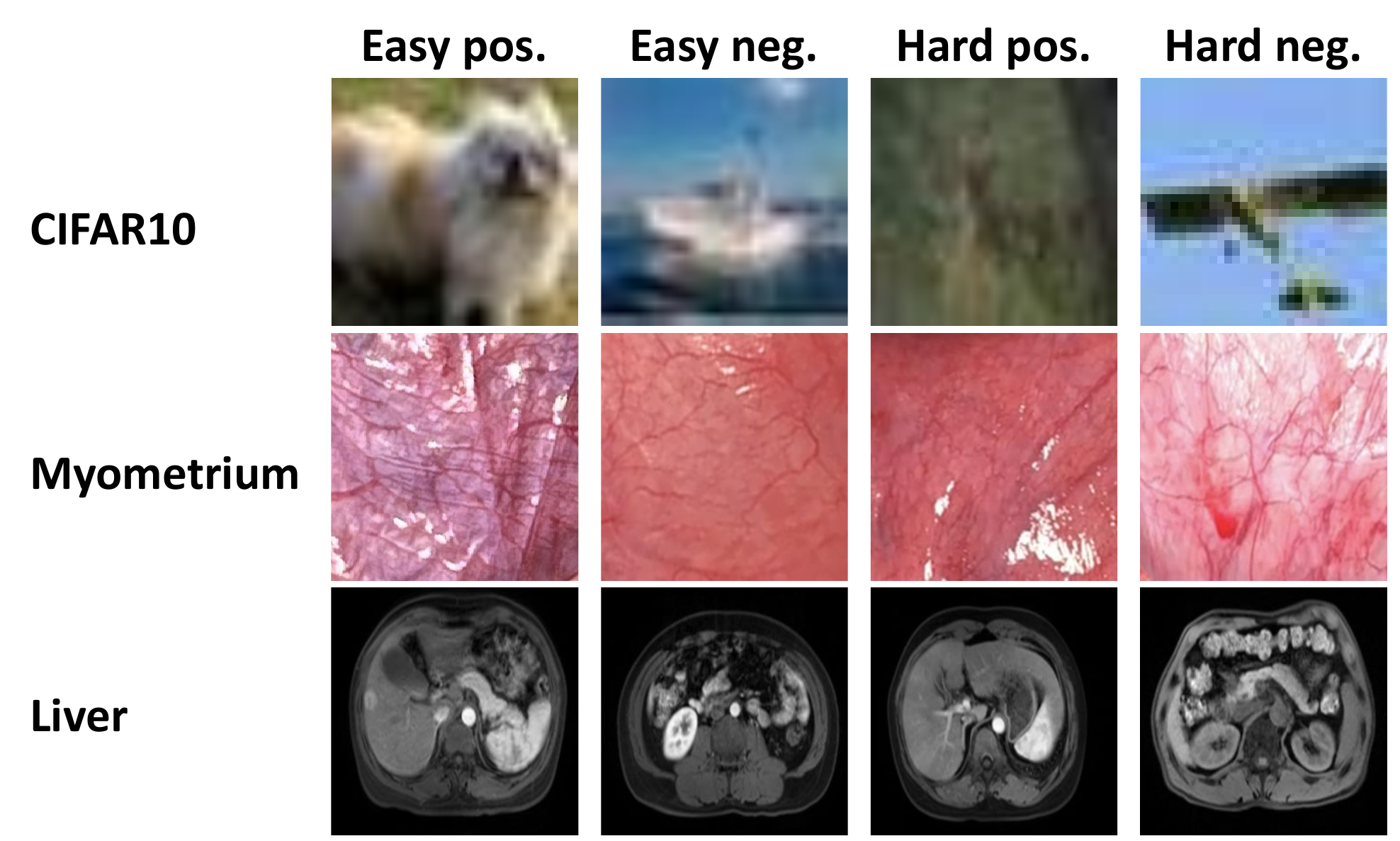}
        \vspace{-0.1in}
    \caption{Visualization of the  easy and hard sets.}
    \label{fig:hard_data}
        \vspace{-0.15in}
\end{figure}
\end{appendices}
\end{document}